# Artificial intelligence system based on multi-value classification of fully connected neural network for construction management

**Tetyana Honcharenko[1,5], Roman Akselrod[2,5], Andrii Shpakov[3,5], Oleksandr Khomenko[4,5]**

[1]Department of Information Technologies of Design and Applied Mathematics, Department of Information Technologies
[2]Vice-rector for Academic Work and Regional Development
[3] Vice-rector for educational and methodical work, Department of Management in Construction
[4] Vice-rector for Academic Affairs and Integrated Development
[5]Kyiv National University of Construction and Architecture, Kyiv, Ukraine

| Article Info | ABSTRACT |
|---|---|
| *Article history:*<br><br>Received month dd, yyyy<br>Revised month dd, yyyy<br>Accepted month dd, yyyy<br><br>*Keywords:*<br><br>Neural network<br>Artificial intelligence<br>Multi-value classification<br>Machine learning<br>Construction management | This study is devoted to solving the problem to determine the professional adaptive capabilities of construction management staff using artificial intelligence systems. It is proposed Fully Connected Feed-Forward Neural Network (FCF-FNN) architecture and performed empirical modeling to create a Data Set. Model of artificial intelligence system allows evaluating the processes in an FCF-FNN during the execution of multi-value classification of professional areas. A method has been developed for the training process of a machine learning model, which reflects the internal connections between the components of an artificial intelligence system that allow it to "learn" from training data. To train the neural network, a data set of 35 input parameters and 29 output parameters was used; the amount of data in the set is 936 data lines. Neural network training occurred in the proportion of 10% and 90%, respectively. Results of this study research can be used to further improve the knowledge and skills necessary for successful professional realization.<br><br> |

*Corresponding Author:*

Tetyana Honcharenko
Department of Information Technologies of Design and Applied Mathematics
Kyiv National University of Construction and Architecture
31, Povitroflotsky Avenue, Kyiv, 03037, Ukraine
Email: iust511@ukr.net

## Introduction

Nowadays Artificial Intelligence becomes one of the defining Start Technologies, which is being worked on in a significant number of areas with regard to the possibility of its implementation.

According to the research with P. Bocher at the front, artificial intelligence is defined as a system that simulates human behavior by analyzing the environment and performing functions, with a certain degree of autonomy, to achieve pre-formed goals [1]. Thus, artificial intelligence is a systemic definition that includes machine and deep learning, peculiarities of which are revealed while artificial intelligence models' learning.

The study [2] provides a scientific basis for the application of artificial intelligence technology to determine the professional adaptive capabilities of construction management staff. If we base the development of an artificial intelligence information system for multi-value classification on the results of youth vocational guidance tests, it will improve the diagnosis of professional selection and will be able to provide recommendations for improving the productivity of professional implementation. The level of trust of young people in the artificial intelligence when it comes to the questions of career guidance, choosing a professional direction or profession according to their wishes, aptitudes, capabilities, and so on, is then determined.

Ukrainian scientists A. Hassan and Y. Palamarchuk analyzed the introduction of artificial intelligence elements in organizing and processing the results of professional orientation. In their work [3], they suggested a structure and methodology for constructing an intellectual system of the vocational guidance for the youth.

Polish scientists P. Szymański, T. Kajdanowicz, and A. Gramfort study the multi-label classification process. The paper [4] presents a multi-value classification for solving several or multiple output values. Scientists note that such a classification should be applied in System cases that have multiple input and output classes, the classification of which, respectively, is necessary.

Scientists from Nigeria A.T. Owoseni, O. Olabode K.G. Akintola, P. A. Enikanselu in [5] established the usefulness of Adaptive neuro-fuzzy inference system in predicting outcomes of events, processes or systems from their numerical datasets. The authors described mathematical details and proposed an improved Adaptive neuro-fuzzy inference system that uses an agglomerative-based clustering ensemble of fuzzy c-means to help extract rules from a given dataset that represent a system.

The work [6] of Turkish scientists N. Acun, E. Ekinci, and S. Omurca states that the practical implementation of the decision tree for machine learning has much more nuances due to the significant number of input and output parameters, which will lead to poor predictability due to weak data normalcy.

In research papers [7, 8, 10] and [11], the authors propose a multi-label classification that predicts the high efficiency of the neural network learning model and the quality of the output classifications in large-scale projects. Accordingly, each multi-value classification task contains its stratification by several parameters and its quality indicators. It also manages a data set with several parameters, performs regression and training by several parameters, and can represent the parameter space [12].

The article [13] suggests that machine learning with a teacher has a certain list of input data that a specialist wishes to use in an artificial intelligence model to denote the initial information. The paper [12] notes that machine learning without a teacher has only input data, where the model forms its structure.

Research papers [14, 15] and [16] explore the regression of machine learning applied in the traditional determination of cause-effect relationships between variables. Thus, the standard category of machine learning with a teacher is a classification task, which is one of the most popular methods for determining cause-effect relationships between variables in the form of a decision tree.

The authors in [17] applied big data technologies, analytics and artificial intelligence to accomplish complex tasks at levels beyond human skills. They proposed the use of artificial intelligence based ensemble learning technique for validating the engineering properties of material.The authors in [18] developed an efficient intrusion model for preventing network intrusion attacks in real-life application scenarios using data analytics and machine intelligence. Few researchers attempted optimizing machine learning algorithms in [19, 20, 21] using open-source systems.

Research papers [22] and [23] suggest to solving the problem to determine the professional adaptive capabilities of construction management based on a multidimensional data model.

The study [24] is devoted to the use of fuzzy logic methods to assess the impact of adaptability factors on the effectiveness of change management at construction industry enterprises.

This study aims to develop a model of artificial intelligence that will solve the problem of multi-label classification of determining the professional adaptive capabilities of construction management staff. To achieve the research goal, it is necessary to investigate aspects of neural networks and their multi-value classification, develop a Fully Connected Feed-Forward Neural Network (FCF-FNN) architecture and perform empirical modeling to create a Data set.

**The main research**

Having analyzed the works and studies of Ukrainian and foreign scientists, we have developed a model of an FCF-FNN to perform the multi-value classification. Fig. 1 shows the architecture of the developed FCF-FNN model.

A neuron consists of synapses that provide communication between the inputs and the neuron kernel, which, in turn, processes input signals and uses the Axon to transmit the processed information to the output of the neuron, that is, to the neuron of another layer.

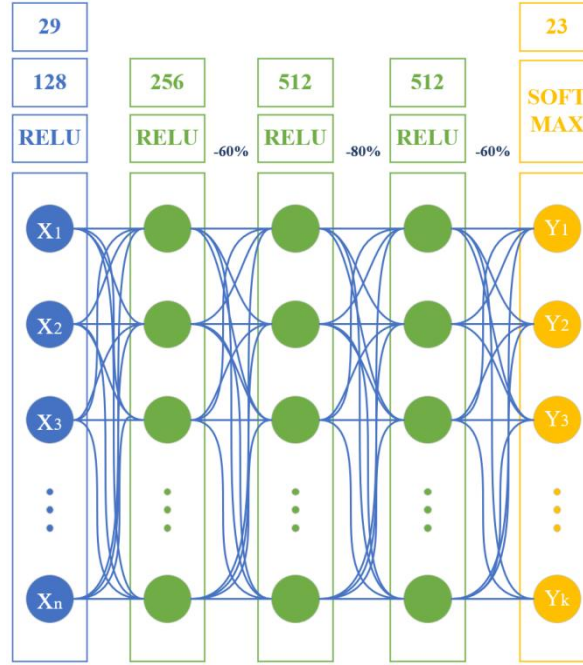

**Fig. 1.** The Architecture of an FCF-FNN model

The significance of input flows is adjusted through synapses that correct the weight that influences the state of the neuron's kernels. Equation for determining the state of the neuron kernel:

$$S = \sum_{i=1}^{n} x_i \cdot \omega_i \qquad (1)$$

where $n$ is number of input neurons;
$x_i$ is value of the corresponding $i^{th}$ neuron;
$\omega_i$ is weight of the corresponding $i^{th}$ synapse.
Equation for determining the value of the output neuron (Axon):

$$Y = F(S) \qquad (2)$$

where $F$ is activation function of input neuron processing (Axon).

The study uses the following activation functions that are suitable for multi-value classification: Rectified Linear Unit (RELU) and Softmax.

The RELU activation function allows for the optimal and productive use of resources displayed in the minimal involvement of neurons that have a simple derivative, where all negative values are 0 and positive 1. Redundant parametrization is a necessary and sufficient condition for achieving zero loss when training a neural network.

Y. Shin and G. Karniadakis recognize one specific issue with using this activation feature, which is called "The Dying RELU". This problem occurs when RELU neurons become inactive and only output the value 0 for any input neuron [13]. This indicates that the neuron is dying, depriving the model of further learning and reducing the accuracy of the classification results. Fig. 2 shows a diagram of the RELU activation function. A ReLU neuron is considered dead if it only outputs a constant for any input. To solve the problem, it is expedient to consider two states of neuronal death — expected and permanent. Then the neural network as a whole will be trainable if the number of permanently dead neurons is small enough for learning.

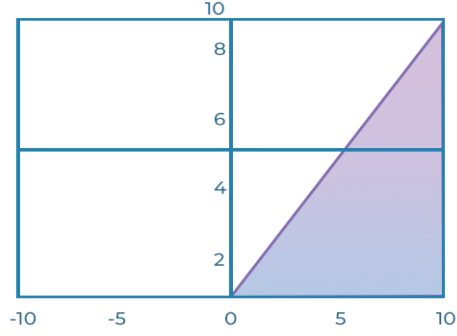

**Fig. 2.** Diagram of the RELU activation function

The Softmax activation function is usually used in the multi-value classification tasks. The advantage of using this function is that the sum of all input neurons will be equal to 1. Thus, the output values for the function are in the range from 0 to 1, which correspond to each output neuron, which is ranked by the probability of output, and whose highest probability, by default, is used in the program. The overall result of this function is a comprehensive prediction of all output neurons, considering the patterns of input neurons. Equation for determining the value of neurons of the softmax group:

$$y_i = \frac{e^{z_j}}{\sum_{i=1}^{K} e^{z_i}} \qquad (3)$$

where $j = 1,…,K$ and determines the number of the output neuron;
$\sum_{i=1}^{K} e^{z_i}$ is adder of all neurons of softmax group;
$e^{z_i}$ is exponent of the output value of the neuron softmax group.

The basis of the neural network model is an FNN multilayer perceptron, one of the basic fundamental architectures of neural networks, occurring when the signals are propagated exclusively from input layers to output. This architecture confirms the complete connection of the neuron to all neurons of the previous layer. Based on the research [9], the chosen architecture is straightforward enough, which is determined by the transfer of information from inputs to outputs without the use of cycles.

The study suggests referring to the classical Titanic problem, which forms a decision tree according to the classification and prediction of smaller subgroups. The work [6] of Turkish scientists N. Acun, E. Ekinci, and S. Omurca states that the practical implementation of the decision tree for machine learning has much more nuances due to the significant number of input and output parameters, which will lead to poor predictability due to weak data normalcy.

Thus, the classic Titanic problem, based on the ticket price, age, class, sex, and several other parameters, allows you to determine the probability of a person's survival. The Problem is solved by predictive analysis. Y. Kakde claims its essence lies in using computational methods that characterize the pattern in large amounts of data [16].

Therefore, it is logical that the prediction of survival in the Titanic problem is performed on different combinations of static features. That is why, to implement artificial intelligence in the organization and process the vocational guidance for secondary school students, the data analysis model is previously considered. This model comprises categories with the input and output data on which the artificial intelligence model will be taught. Fig. 3 shows the model and categories of artificial intelligence data analysis.

Accordingly, a quantitative sample of respondents sets a threshold value for a set of tests for training the model. Input and output descriptions are normalized and partially presented in Table 1. It is worth noting that the list of professional areas has a primary source – the National Career system of Great Britain, which was significantly adapted for the study needs.

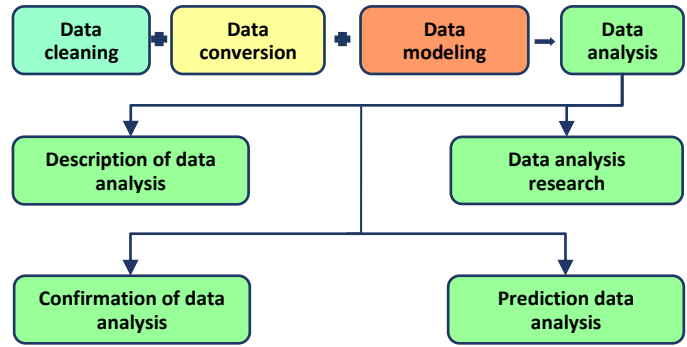

**Fig.3.** Model and categories of artificial intelligence data analysis research

**Table 1**
Description and characteristics of machine learning model data

| № data parameter | Name of the data parameter | Description | Characteristics |
|---|---|---|---|
| 1 | Age | Respondent's age (full years) | $0 \leq \frac{Age}{\max\_Age} \leq 1$, when $\max\_Age$ corresponds to the value of the maximum age of respondents |
| 10 | AT | Percentage of self-confidence that takes the maximum value of 100 | $0 \leq \frac{AT}{\max\_q} \leq 1$, when $\max\_q$ corresponds to the maximum value of 100 |
| 11 | TT2 | Percentage of caution that takes the maximum value of 100 | $0 \leq \frac{TT2}{\max\_q} \leq 1$, when $\max\_q$ corresponds to the maximum value of 100 |
| 12 | RPT | Indicator of a realistic professional personality type that takes the maximum value of 14 | $0 \leq \frac{RPT}{\max\_PT} \leq 1$, when $\max\_PT$ corresponds to the maximum value of 14 |
| 13 | IPT | Indicator of an intellectual professional personality type that takes the maximum value of 14 | $0 \leq \frac{IPT}{\max\_PT} \leq 1$, when $\max\_PT$ corresponds to the maximum value of 14 |
| 14 | APT | Indicator of an artistic professional personality type that takes the maximum value of 14 | $0 \leq \frac{APT}{\max\_PT} \leq 1$, when $\max\_PT$ corresponds to the maximum value of 14 |
| 32 | CVW | Normalized indicator of the professional direction "Charity and voluntary work" | CVW |
| 35 | EM | Normalized indicator of the professional direction "Engineering and manufacturing" | EM |
| 37 | H | Normalized indicator of the professional direction "Healthcare" | H |
| 50 | SC | Normalized indicator of the professional direction "Social care" | SC |

As part of the study, the training Data set is formed by a comprehensive survey. To train, the model uses the results of a survey. The respondents have already worked in a particular professional field they find interesting. Thus, an empirical study reveals the normalizability of the sample for training the machine learning model, as well as for further classification and forecasting the outcomes for those respondents who have not yet decided on a professional direction. The information about them will also be obtained from the survey.

Fig. 4 shows an algorithm for the training process of a machine learning model, which reflects the internal connections between the components of an artificial intelligence system that allow it to "learn" from training data.

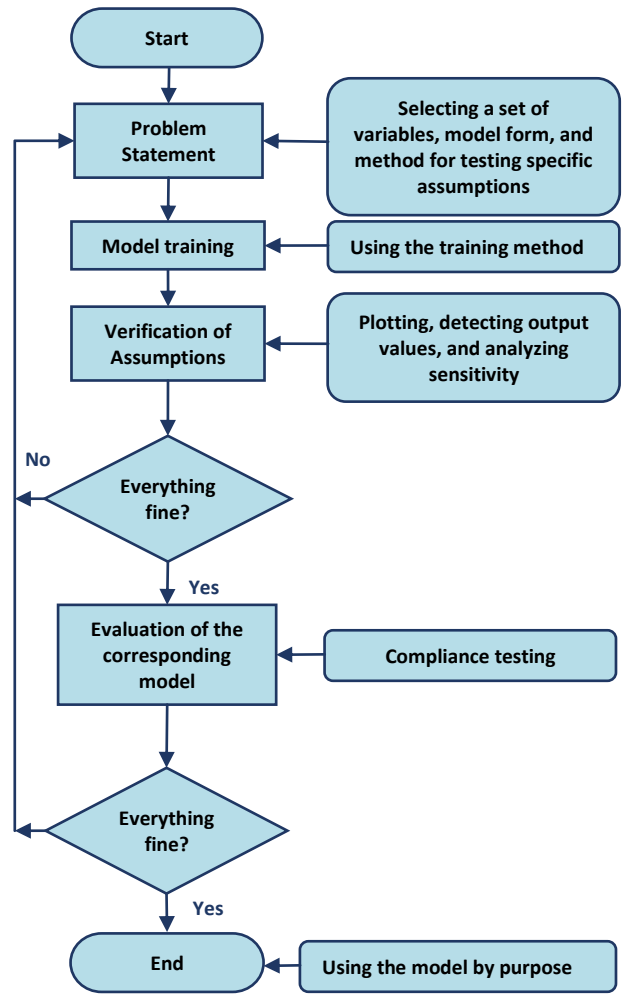

**Fig. 4.** Algorithm of the machine learning model training

A functional model based on the IDEF (Integration Definition for Function) approach has been developed to describe the operation of the information system and the processes taking place within it. Fig. 5 shows diagram for Model of Fully Connected Feed-Forward Neural Network, which allows to evaluate the processes occurring in an FCF-FNN during the execution of multi-value classification.

The FCF-FNN model has a comprehensive description of the information system. Input arrow "SCV file" is a set of information type int64 about a person undergoing professional orientation.

The output data of the system is:

1) An SCV file containing add-ons by the performed multi-value classification by professional areas and can be exported for use in other information systems;

2) DataFrame is a table that saves the data and allows for viewing processed data and data for further processing;

3) An H5 file is a file of an artificial intelligence model that has undergone appropriate training and can be exported for implementation to another information system.

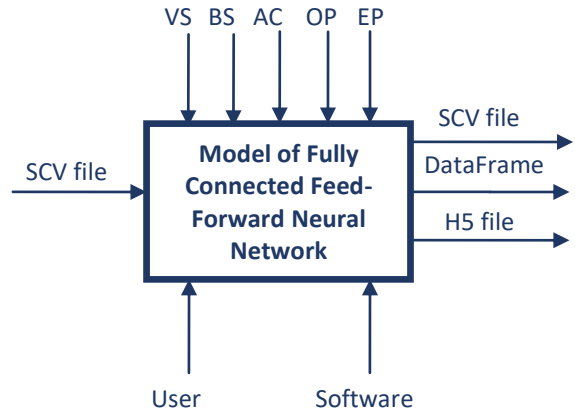

**Fig. 5.** Model of Fully Connected Feed-Forward Neural Network

The "User" and "Software" arrow define the data transformation mechanisms (signal system guidance elements). The arrows VS (validation_split), BS (batch_size), AC (activation), OP (optimizer), EP (Epochs) are the tools for implementing the data transformation. VS determines how to regulate the percentage of training data to test them. BS determines the number of rows of data that will be fed to the model during one epoch. AC (activation function) defines the type and manner of numeric data processing. OP determines the optimization algorithm by which the model will work. EP defines the number of repeated passes of study on training Data sets.

The software product was developed by Python and uses Keras, Numpy and Pandas libraries. Explaining research chronological, including research design, research procedure (in the form of algorithms, Pseudocode or other).

**Results**

To create a Data Set for subsequent training, we used 35 input parameters and 29 output parameters obtained empirically by research input and multiplication, preserving the mathematical median of the results. The amount of data in the set equals 936 rows of data. Neural network training occurred in the proportion of 10% and 90%, respectively.

The total training time was approximately 30 minutes, with 2 approaches for 100 epochs and 1 for 1000 epochs. The average duration of one epoch is 1 second of real-time, with 20 lines of data per learning epoch. According to the results, it is logical that the model has already finished learning by the 300th general epoch. However, an additional overtraining check was performed. As a result, the model did not begin to degrade rapidly over 700 overtime periods and was able to maintain the accuracy of the answers.

Fig. 6 shows a success of training the model.

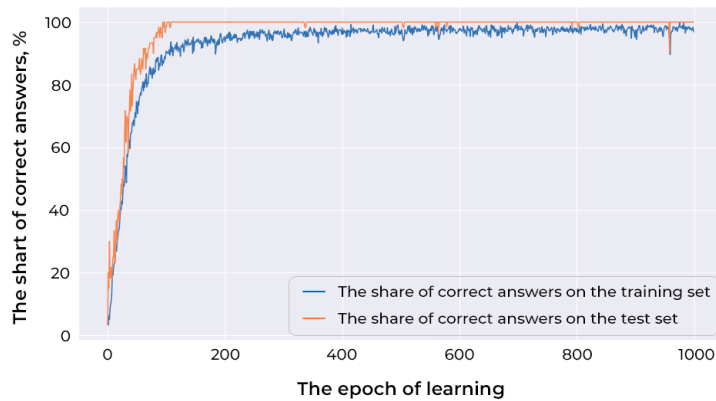

**Fig. 6.** Diagram of the correct answers ratio on training and test sets for a period of 1000 training epochs

Fig. 7 shows a mistake-ratio curve on the training and test sets, which means the neural network works correctly, minimizing the number of mistakes after sufficient learning time.

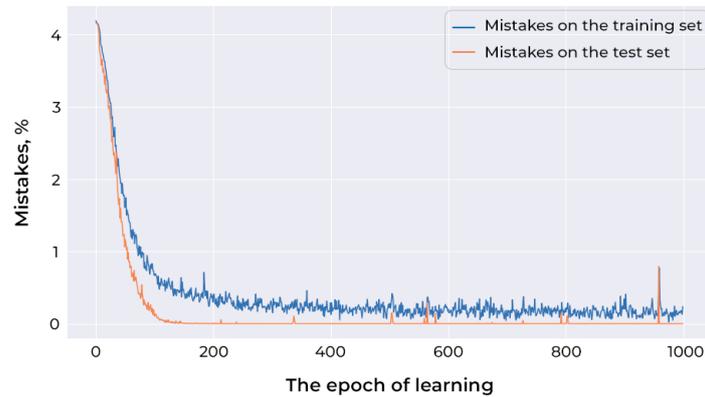

**Fig.7.** Diagram of the mistake ratio for a period of 1000 epochs

According to the empirical study data, a number of real data of respondents was obtained, which made it possible to find out their compliance with a certain professional profile. Table 2 shows two popular results, namely respondents № 10 and № 35, with the following probabilities:

**Table 2**
Interpretation of neural networks results

| Respondent's identification № | Interpretation of neural networks results | |
|---|---|---|
| | *Professional direction profile* | *Probability of successful professional implementation* |
| 10 | EA | 94,6% |
| | EU | 3,5% |
| | EM | 0,9% |
| 35 | CVW | 99,9% |
| | SC | 0,0000033% |

The overall results of each respondent of the study have an ultra-high accuracy of giving priority professional directions because the system performs ranking according to the highest compliance with the values of 1. Respectively, "weak" by nature professional directions can have the value "number · $e^{-19}$". However, in the cases similar to respondent's № 35, it is worth considering even "weak" manifestations of professional trends, because an alternative allows the individual to better understand themselves.

The respondents who are still looking for a job mainly demonstrated that the profession they learned (the first-course students were the main participants) coincided with the results by artificial intelligence.

According to the study carried out, we consider it appropriate to recommend involving a developed artificial intelligence system to offering career guidance to the future university students. Besides, our research results can be used to further improve the knowledge and skills necessary for successful professional realization.

**Conclusion**

The study proposes a solution to the problem of multi-value classification for determining the professional adaptive capabilities of construction management staff. The performed empirical modeling for creating a dataset confirms the adequacy of the FCF-FNN model to solve the problem. The proposed neural network architecture is straightforward, defined by the coherent transmission of information from inputs to outputs without the use of loops. When designing the Model, 2 activation functions were used: RELU and Softmax, which made it possible to rank the results of a multi-valued classification of professional areas by priorities in the range of values from 0 to 1. The constructed model contains 128 Dense input layers (35 input parameters), 1256 hidden Dense layers, with an arbitrary dropout sample reduction of 60%, 80%, 60%, and then the data are sent to 29 output neurons of the softmax group.

The authors believe it is possible to develop a neural network model for the formation of a national database that will store "digital portraits of the individual", according to which employers will be able to set specific requirements for staff of construction management. Such "digital portraits of the staff " can serve as both a virtual summary and a portfolio at the same time.

**Author's contribution**

All authors contributed equally towards the manuscript concept formulation, writing and results analysis.

**Availability of data and materials**

Data sharing is not applicable to this article as no datasets were generated during the current study. All data generated or analyzed during this study are included in this article.

**Declaration of Competing Interest**

The authors declare that they have no competing interests.

**Funding**

This study received no external funding.